\documentclass[letterpaper, 10 pt, conference]{ieeeconf}  

\IEEEoverridecommandlockouts                              

\usepackage{amsmath} 
\usepackage{amssymb}  
\usepackage{graphicx}
\usepackage{wrapfig}
\usepackage{adjustbox}
\usepackage{makecell}
\usepackage{multirow}
\usepackage{float}
\usepackage{cite}
\usepackage[dvipsnames]{xcolor}
\usepackage{caption}
\usepackage{threeparttable}
\usepackage{url}

\title{\LARGE \bf
A Collaborative Team of UAV-Hexapod for an Autonomous Retrieval System in GNSS-Denied Maritime Environments
}

\author{Seungwook Lee$^{+}$, Maulana Bisyir Azhari$^{+}$, Gyuree Kang, Ozan Günes, \\
Donghun Han, and David Hyunchul Shim$^{*}$
\thanks{$^{+}$Equal Contributions}
\thanks{$^{*}$Coressponding author}
\thanks{All authors are with the School of Electrical Engineering, Korea Advanced Institute of Science and Technology, Daejeon, South Korea
        {\tt\small \{seungwook1024, mbazhari, fingb20, ozan.guenes, donghunhan, geninfty\} @kaist.ac.kr}}%
}

\begin{document}

\maketitle
\thispagestyle{empty}
\pagestyle{empty}

\begin{abstract}

We present an integrated UAV-hexapod robotic system designed for GNSS-denied maritime operations, capable of autonomous deployment and retrieval of a hexapod robot via a winch mechanism installed on a UAV. This system is intended to address the challenges of localization, control, and mobility in dynamic maritime environments. Our solution leverages sensor fusion techniques, combining optical flow, LiDAR, and depth data for precise localization. Experimental results demonstrate the effectiveness of this system in real-world scenarios, validating its performance during field tests in both controlled and operational conditions in the MBZIRC 2023 Maritime Challenge.


\end{abstract}

\section{INTRODUCTION}

Unmanned Aerial Vehicles (UAVs) have become an essential component of modern robotics, widely used in various applications, including surveillance, inspection, search and rescue, and transportation. Their ability to fly over challenging terrains and access remote areas has expanded the scope of autonomous operations in many sectors. However, UAVs face limitations when it comes to direct interaction with the environment, particularly when ground-based tasks are required. On the other hand, ground robots, such as hexapods, are well-suited for traversing uneven and complex terrains, offering a high degree of stability and adaptability. These ground robots excel in direct manipulation tasks, such as object retrieval, inspection, and repairs, making them ideal for environments where UAVs alone cannot perform optimally in.

\begin{figure}[t!]
\centering
    \includegraphics[width=0.95\columnwidth]{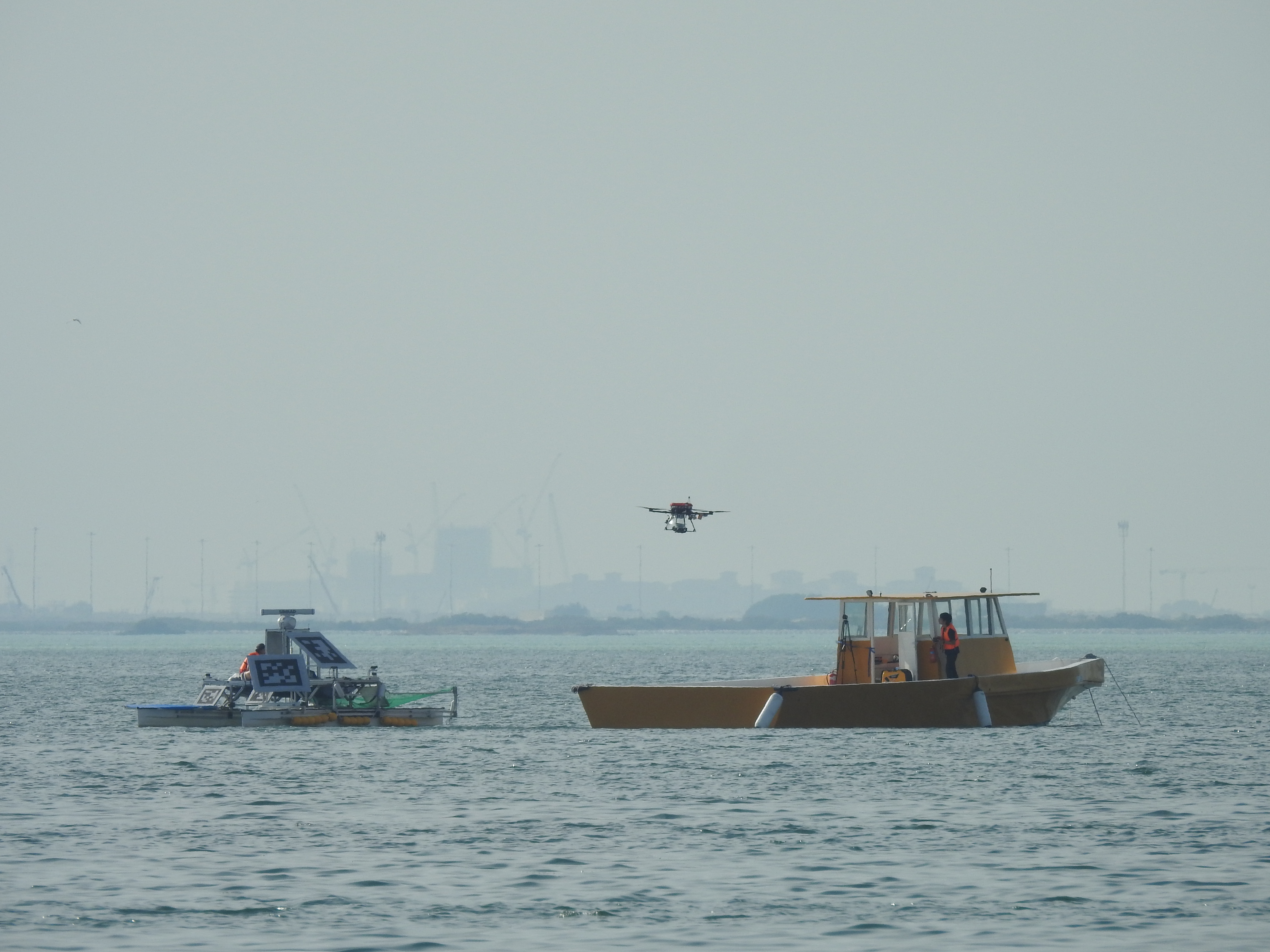}
    \caption{UAV-Hexapod system executing its mission in a GNSS-denied maritime environment. Team KAIST won 2nd place in the MBZIRC 2023 Maritime Challenge. }
    \label{fig:cover_picture}
\end{figure}

The integration of UAVs and ground robots into a unified system presents an effective approach to object retrieval in challenging maritime environments. Maritime environments are distinguished by dynamic conditions, including wave movements, the lack of fixed landmarks, and adverse weather. In these settings, the combination of aerial and ground-based capabilities enables more effective and flexible mission execution. For instance, a UAV can provide long-range navigation and rapid deployment capabilities, while a ground robot can execute tasks that require stability and precise interaction with objects or moving surfaces.

However, operating in GNSS-denied maritime environments introduces several challenges that traditional UAV systems cannot easily overcome. One of the primary difficulties is reliable localization. In settings such as the MBZIRC 2023 Maritime Challenge where GPS signals are unavailable, traditional localization methods that rely on visual data are prone to fail. Also, the dynamic nature of the maritime environment, including moving surfaces like waves, complicates the UAV's ability to maintain accurate positioning which becomes very important when it comes to object retrieval tasks. Weather conditions act adversely as well. Extreme lighting conditions, such as overexposure due to direct sunlight or underexposure in backlit conditions, further degrade the performance of vision-based localization systems. Strong wind around the coastal area requires sufficient wind resistance for the UAV platforms.

This paper presents the design, implementation, and experimental validation of the integrated UAV-Hexapod system, focusing on its ability to perform tasks in real-world maritime environments. The system's performance is evaluated in the context of the MBZIRC 2023 Maritime Challenge, where it demonstrates its capability to tackle dynamic and complex scenarios. The rest of the paper is organized as follows: Section \ref{sec:related_work} presents prior works, Section \ref{sec:system_design} details the system design, Section \ref{sec:method} discusses the localization methods and hexapod operation, Section \ref{sec:experiments} presents the experimental results, and Section \ref{sec:conclusion} concludes with insights and future directions for improving the system.

\section{Related Work}
\label{sec:related_work}
\subsection{GNSS-denied Localization for UAV}
GNSS-denied UAV navigation has gained significant attention due to the limitations of GNSS-based systems, which are vulnerable to jamming, signal loss, and environmental obstructions \cite{gyagenda2022gnssreview}. 
In GNSS-denied environments, UAV localization relies on alternative methods such as visual odometry, LiDAR, and inertial navigation. 
Visual odometry is widely used to estimate UAV motion, performing well in many environments but struggling in low-texture or dynamic scenes \cite{mur2015orb1, mur2017orb2, campos2021orb3, forster2016svo, forster2014svo1, engel2017dso}. 
LiDAR-based localization excels in feature-rich areas, generating detailed maps for precise localization through scan-matching and pose graph optimization, though its effectiveness is limited in open spaces like oceans ~\cite{zhang2014loam,guo2022loam, shan2018lego, zhang2017low}. 
Recent approaches, like fiducial markers (e.g., AprilTag), provide reliable reference points, especially in GNSS-denied environments, but are constrained by camera range, field of view, and lighting conditions \cite{wang2016apriltag, olson2011apriltag, 
fiala2009artag, garrido2014aruco, 
aiaascitech2024multiscale_fiducial, bauschmann2023evaluation_underwater_apriltag}. 
Optical flow-based localization offers another solution but depends on accurate height estimation and struggles in featureless environments like water \cite{hou2019flow2, zhang2016flow1, mebarki2014flow3}.
However, in the maritime environment for MBZIRC 2023 challenge, it is difficult to properly localize the UAV, since there is no sufficient geometric or visual features. We explore the possibility of utilizing fiducial markers, optical flow, and various height measurement sensors for localizing the UAV in the maritime environment.

\subsection{Object Retrieval}

Aerial manipulation is a rapidly advancing field that combines the mobility of UAVs with the dexterity of robotic manipulators, enabling UAVs to engage in complex tasks requiring direct interaction with objects. Initial developments in this area focused on integrating multi-degree-of-freedom (DOF) robotic arms with UAVs to perform tasks such as grasping and object transport~\cite{orsag2017dexterous, garimella2015mpc_pick_place}. However, these early systems faced significant challenges, particularly due to the dynamic coupling between the UAV and the manipulator, which impacted flight stability and control \cite{li2023nonlinearmpc}. Additionally, when UAVs are positioned too close to an object, they may lose visibility, forcing them to rely on blind estimation or employ repeated attempts to locate the object. This problem is especially pronounced in visual-servoing-based aerial manipulation which relies on the visibility of the object~\cite{he2023servo1, ye2023visionservo2, chen2022imageservo3}.

The object retrieval task involving a UAV can be interpreted as a subset of aerial manipulation. Our approach mitigates these issues by using a winch mechanism to deploy a hexapod robot onto the ground to grasp the object, thus minimizing the time during which the UAV experiences dynamic coupling with the manipulator. Once the hexapod is deployed, the UAV is freed from the destabilizing effects of direct manipulation, allowing it to maintain a stable flight and a clear field of view around the object of interest. This solution not only enhances stability but also improves the UAV’s situational awareness, enabling more precise and efficient object retrieval.




\begin{figure*}[t]
    \centering
    \includegraphics[width=0.95\textwidth]{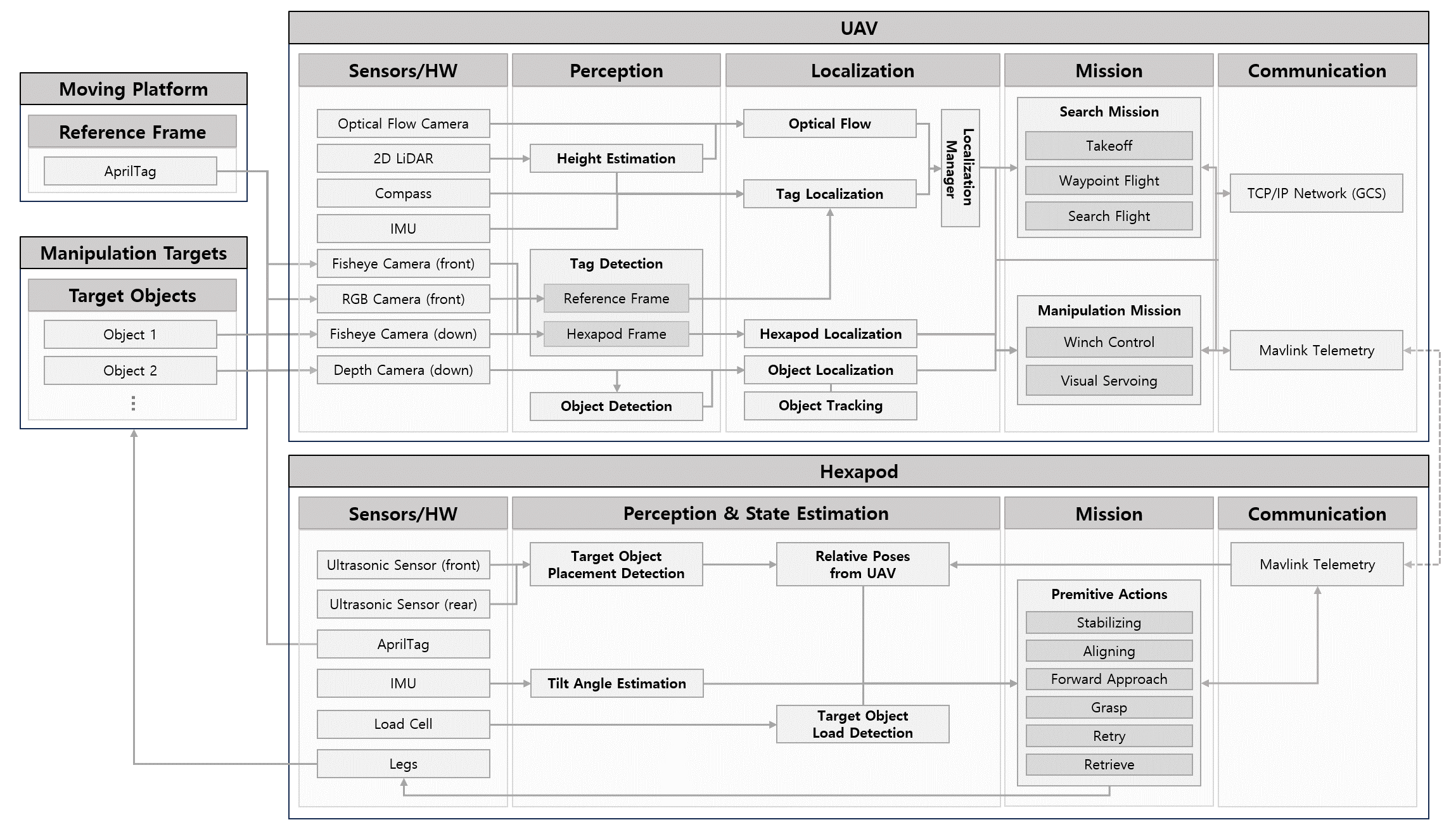}
    \caption{Overview of the proposed UAV-Hexapod system for object retrieval tasks in GNSS-denied maritime environment}
    \label{fig:system_diagram_all}
\end{figure*}

\section{SYSTEM DESIGN}
\label{sec:system_design}
The proposed UAV-Hexapod system, described in Fig. \ref{fig:system_diagram_all} is designed to address the challenges of maritime object retrieval in a GNSS-denied environment. The system consists of two main components: a UAV and a hexapod, which are connected via a winch mechanism. The system allows effective and reliable object retrieval on unstable and dynamic surfaces.

\subsection{UAV}
The UAV serves as the primary transport and reconnaissance platform. It is equipped with a range of sensors that enable stable flight and localization in the GNSS-denied environment. The main sensors include the following: two T265 fisheye cameras for localizing the UAV relative to a set of fiducial markers installed on a reference frame as the primary source of localization of the UAV; A D455 depth camera for close-range measurements directly over the target and secondary height estimation; an optical-flow sensor as a secondary localization modality; and A Hokuyo 2D LiDAR for primary height estimation over water, providing distance measurement even in the absence of visual features. The optical-flow data is fused with the 2D LiDAR and depth image for accurate localization over water. This sensor fusion approach compensates for the lack of GNSS signals and unreliable visual cues often encountered in the maritime setting. 

A winch system is mounted on the UAV, enabling controlled deployment and retrieval of the hexapod robot. This winch operates in a closed-loop system with feedback from the ultrasonic sensor installed on the UAV and a small set of fiducial markers installed on the hexapod to ensure safe and accurate lowering and retrieval.

\subsection{Hexapod}
The hexapod is designed for stable object retrieval on dynamic surfaces, such as vessels. It features six legs that provide stability and adaptability on uneven terrain. The robot’s legs also serve as manipulators, allowing it to grasp objects efficiently. Once it grabs the object, it is pulled with the object to the desired location by the UAV. The hexapod is equipped with a lightweight frame made from magnesium alloy and 3D-printed components to reduce its overall weight. Its onboard computer, an Nvidia Jetson Xavier NX, handles sensor data processing, communication with the UAV, and control of gait and grasping functions.

The hexapod’s key features include: Autonomous docking and undocking with the UAV via the winch mechanism; adaptive gait control, enabling it to maintain balance on moving platforms; and ultrasonic sensors for object detection and distance measurement, which aid in identifying and grasping the target once it is guided to the area by the UAV.



\subsection{Mission Planning}
The mission planning process is driven by the UAV’s flight status and detection information. Upon receiving a target location or command, the UAV navigates to the area discovered during the search mission using its onboard sensors. Once the UAV has positioned itself above the target area, the hexapod is deployed via the winch mechanism.

The hexapod’s mission planning is based on sensor input and detection information from the UAV. Upon deployment, it heads towards the target object using the relative target location provided by the UAV. Once it arrives over the target object, onboard sensors detect the object and the robot applies a firm grasp with its legs.

\subsection{Communication}
Communication between the UAV and the hexapod is achieved through wireless RF communication using a custom MAVLink protocol. This protocol enables the exchange of mission states and sensor data. The UAV and hexapod maintain a continuous data link during operations, ensuring real-time coordination between the two platforms.

The UAV acts as the central node, processing data from its own sensors including the hexapod’s pose, allowing it to adjust its flight path or deployment strategy based on the hexapod’s progress.



\begin{figure*}[t!]
\centering
\includegraphics[width=0.8\textwidth]{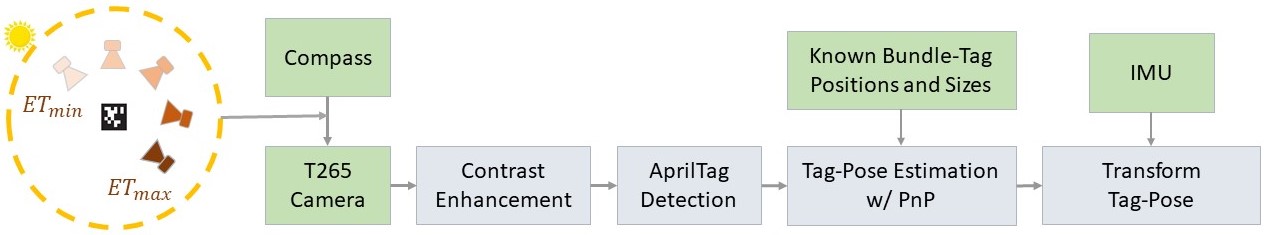}
\caption{The Proposed AprilTag-based Relative Localization Pipeline}
\label{fig:apriltag_loc_pipeline}
\end{figure*}

\section{METHOD}
\label{sec:method}
\subsection{UAV Localization}
In the GNSS-denied maritime environment, we used AprilTag markers for UAV localization relative to the fixed frame on the USV. Detection was affected by the adverse lighting, so camera exposure was adjusted using the UAV's heading and sun position, enhanced by gamma correction. During object retrieval, we employed optical-flow-based localization to stabilize the reference frame relative to the object and used multiple height measurement modalities for robustness. Optical flow can also serve as a fallback when AprilTag detection fails. The UAV localization system is integrated with the task planner to select the best method for each stage of the mission.

\subsubsection{AprilTag-based Relative Localization}
Two on-board fisheye cameras are utilized to capture the surroundings of the UAV, one facing front and one facing down. 
A detection algorithm \cite{kallwies2020determiningapriltag} extracts AprilTags from the image and computes their relative position to the camera. 
Let $x_{UAV}^\mathcal{W} = [p_{UAV}^\mathcal{W}, q_{UAV}^\mathcal{W}]^T$ be the pose of the body frame $\mathcal{B}$ UAV in world coordinate frame $\mathcal{W}$, where $p_{UAV}^\mathcal{W}=[x, y, z]^T$ is the position of the UAV and $q_{UAV}^\mathcal{W} = [\phi, \theta, \psi]^T$ is the orientation containing the roll $\phi$, pitch $\theta$ and yaw $\psi$ angles of the UAV. 
While the UAV's orientation can be estimated by using the IMU, the UAV's position is estimated by the inverse of the tag relative pose such that $x_{UAV}^\mathcal{W}=q_{UAV}^\mathcal{W}(-p_{tag}^\mathcal{B}) - p_{tag}^\mathcal{B}$ where $p_{tag}^\mathcal{B}$ is the detected position of the tag in body frame $\mathcal{B}$. 



AprilTag detection is challenging in outdoor lighting due to the automatic exposure controller's limitations. We propose a heading-aware exposure controller that adjusts based on the sun's position, as depicted in Fig.~\ref{fig:apriltag_loc_pipeline}. Assuming the UAV always faces the AprilTag, we approximate its direction as opposite to the UAV heading.
With a known sun's direction, we can exploit the information for setting the appropriate exposure time $t_{exp}$
\begin{equation}
    t_{exp}=\frac{1}{2}(\cos(\psi_{sun} - \psi_{uav})+1)(t_{exp,max}-t_{exp,min}) + t_{exp,min}
\label{eq:heading_aware}
\end{equation}
where $\psi_{sun}$ is the sun position in the sky in term of heading direction, $\psi_{uav}$ is the UAV's heading, $t_{exp,min}$ and $t_{exp,max}$ are the minimum and maximum exposure time obtained by the exposure calibration before the deployment.

In addition to the heading-aware exposure controller, we apply a contrast enhancement technique to improve AprilTag detection. Specifically, we use adaptive gamma correction with weighting distribution (AGCWD) from \cite{huang2012agcwd}, which adjusts intensity using the transformation $l^{'}=l_{max}(l/l_{max})^\gamma$, where $\gamma$ is computed using $1-cdf_w(l)$ based on the cumulative distribution function $cdf_w$ of pixel intensity $I$.

\begin{equation}
    cdf_w(l) = \sum_{l=0}^{l_{max}}pdf_w(l)/\Sigma pdf_w
\end{equation}
\begin{equation}
    pdf_w(l) = pdf_{max}\left(\frac{pdf(l)-pdf_{min}}{pdf_{max}-pdf_{min}}\right)^\alpha
\end{equation}
where $pdf_w(l)$ is probability density function parameterized by $\alpha$, $pdf_{min}$ and $pdf_{max}$ are the minimum and maximum $pdf$ of the statistical histogram.

\subsubsection{Optical-Flow-based Localization}

Once the target object is detected during retrieval, following the USV for AprilTag-based localization is unnecessary and problematic due to the moving origin point. At this point, we assume the vessel is directly below the UAV, allowing us to switch to optical-flow-based localization using the vessel's visual features. The optical-flow system outputs the UAV's world-frame velocity $v_{UAV}^\mathcal{W}$, which the localization manager uses to determine the UAV's 3D pose.


We measure the optical flow of the ground beneath the UAV using the UPFlow LC302, which provides the perceived rotational velocities. They are transformed into the world frame using height measurements and IMU angular velocities.

Height measurement poses a significant challenge for optical flow in maritime environments. Most drone altimeters, such as barometers, sonar sensors, or laser-based systems, fail to provide the necessary accuracy and range for this application. As alternatives, we use two primary height sensors: the Intel Realsense D455 depth camera and the Hokuyo 2D LiDAR. The D455 measures height when the UAV is above the vessel, with a maximum range of 8 meters, while the Hokuyo 2D LiDAR provides height measurements over water surfaces, with a detection range of up to 20 meters.

\subsubsection{UAV Localization Manager}
To seamlessly switch between localization sources, the reference frames need to be managed accordingly by the localization manager. It has three main purposes. First, it is coupled with the task planner to choose the appropriate localization modality depending on the task stage. Second, when the tag detection is unstable or unavailable, it switches the modality into the optical flow. Lastly, every time the localization modality is switched, the localization manager will update the origin of the reference frame, thus the pose changes are kept continuous and stable for the EKF fusion in the UAV state estimation. 



\subsection{Object Detection and Tracking}
We utilize YOLOv7-tiny\cite{wang2023yolov7} model with TensorRT inference machine, as it is lightweight and compatible with limited onboard computation resources. YOLOv7-tiny detects objects in the image, resulting in 2D detections that show the positions of these objects in the 2D image plane. Afterward, we project the 2D detections on the image plane and then onto the 3D body frame coordinate utilizing the depth image captured by the depth camera.

To handle multiple objects in the scene and possibly false positive detections, we employ a tracking algorithm, namely SORT(Simple Online and Realtime Tracking)~\cite{bewley2016sort}. It tracks the 3D detections across different frames, ensuring that the identity of each object is maintained over time. This pipeline effectively combines object detection, depth information, and tracking to enable real-time 3D tracking of objects.

\label{sec:experiments}
\begin{figure*}[t]
    \centering
    \includegraphics[width=0.90\textwidth]{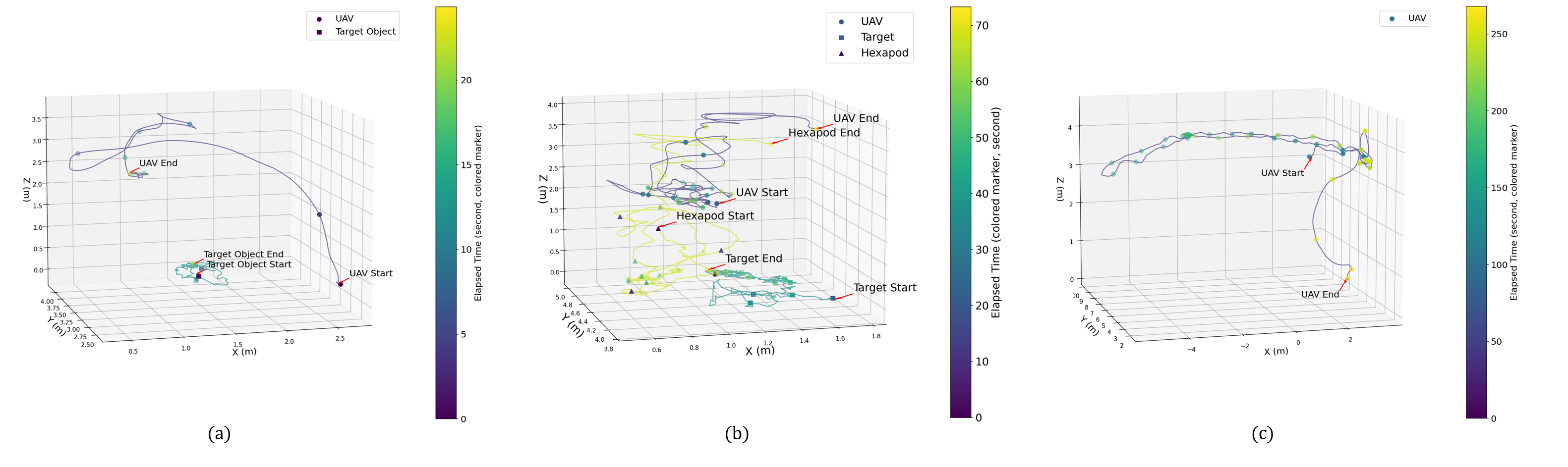}
    \caption{UAV-Hexapod system executing the object retrieval task. (a) UAV takes off and searches for the target object. (b) Hexpod is deployed from UAV and performs retrieval tasks. (c) UAV navigates to a waypoint to complete the retrieval task and returns to base. The entire task is completed in 4 minutes and 27 seconds.}
    \label{fig:3d_plot}
\end{figure*}

\subsection{UAV-Hexapod Tethered System Configuration}
For object retrieval, mobility, and secure grasping are crucial for the hexapod. With its six legs, it ensures stable movement across various terrains, even when some legs encounter obstacles. This multi-legged design provides reliable balance, essential for dynamic environments. Additionally, the hexapod’s legs function as manipulators, enabling stable object grasping as if grasping with a six-fingered hand, accommodating various object shapes and orientations.

\subsection{Hexapod System for Stable Grasping}
The hexapod uses a combination of ultrasonic and load sensors to detect and center on the target. Two ultrasonic sensors, positioned at the front and rear, ensure the target is within the robot's grasping range, providing redundancy in case of sensor failure. Load sensors confirm the target's stability post-grasping, both when grounded and airborne, enhancing robustness.

The onboard IMU tracks the hexapod's orientation by monitoring its tilt relative to the gravity vector. In case of instability, the system triggers a retry mechanism to adjust the robot's posture, ensuring the mission continues safely.

\subsection{Robust Action Selection in Dynamic Environments}
Conventional approaches, such as applying a predefined sequence of actions, may often fail in dynamic conditions that require flexible decision-making in real-time. To achieve this, we use \textit{goal proximity weight} and \textit{action affordance} metrics to select the most suitable actions dynamically.
\[
P = \arg\max_{i \in N} (g_i \cdot a_i)
\]

Goal proximity weight \(g_i\) increases the priority of actions that bring the hexapod closer to the mission objective where \(N\) is the number of primitive actions. Action affordance \(a_i\) evaluates the feasibility of actions based on the current state. The action with the highest combined priority \(P\) is selected at each moment.

\section{Experiments}


To validate the effectiveness and efficacy of the proposed method, we perform field tests during our preparation for the MBZIRC Competition. First, we validate the relative localization system in Port Jongguk, South Korea. Second, we validated our proposed system during the experiment session of MBZIRC competition in Abu Dhabi, United Arab Emirates.

The deployment and retrieval of the hexapod robot were tested in both controlled and real-world environments, including on the deck of a moving vessel. 
Field tests were also conducted during the MBZIRC competition, where the system was deployed in a real-world scenario. The UAV-hexapod system demonstrated its ability to perform complex tasks in challenging maritime conditions, validating the practicality of the proposed approach.

\section{CONCLUSIONS} 
\label{sec:conclusion}
This paper presents a novel UAV-Hexapod system designed for object retrieval in maritime environments. The integrated system addresses the key challenges, including robust localization over water in a GNSS-denied environment, and efficient deployment of a hexapod robot. Experimental results demonstrated the effectiveness of the proposed system in real-world environments, presenting a promising new paradigm for aerial manipulation using unmanned aerial vehicles even in dynamic scenes.
\addtolength{\textheight}{-12cm}   






\section*{ACKNOWLEDGMENT}

The research is supported by Intelligent Countermeasure Technology Development Project for Illegal Drones through the National Research Foundation of Korea(NRF) funded by the Ministry of Science, ICT (2021M3C1C4085706).


\bibliographystyle{ieeetr}
\bibliography{refs}

                                  
\end{document}